\documentclass[lettersize,journal]{IEEEtran}
\usepackage{amsmath,amsfonts}
\usepackage{algorithm}
\usepackage{algpseudocode}
\usepackage{array}
\usepackage[caption=false,font=normalsize,labelfont=sf,textfont=sf]{subfig}
\usepackage{textcomp}
\usepackage{stfloats}
\usepackage{url}
\usepackage{verbatim}
\usepackage{graphicx}
\usepackage{cite}
\begin{document}

\title{GARA: A novel approach to Improve Genetic Algorithms' Accuracy and Efficiency by Utilizing Relationships among Genes}

\author{Zhaoning Shi\textsuperscript{1,*}, Mengxiang\textsuperscript{2,*}, Zhaoyang Hai\textsuperscript{1}, Xiabi Liu\textsuperscript{1,$\dag$}, Yan Pei\textsuperscript{2,$\dag$}

\thanks{\textsuperscript{*}: Equal contributions}

\thanks{\textsuperscript{$\dag$}: Corresponding author }

\thanks{\textsuperscript{1} Z. Shi, Z. Hai and X. Liu are with School of Computer Science and Technology, Beijing Institute of Technology, Beijing, China, 100081.  
(e-mail:\{3220235133, haizhaoyang, liuxiabi\}@bit.edu.cn)}

\thanks{\textsuperscript{2} X. Meng and Y. Pei are with School of Computer Science and Engineering, University of Aizu, Aizuwakamatsu, Japan, 965-8580.
(e-mail:\{d8242104, peiyan\}@u-aizu.ac.jp)}

\thanks{Manuscript received April 19, 2021; revised August 16, 2021.}}

\markboth{Journal of \LaTeX\ Class Files,~Vol.~14, No.~8, August~2021}%
{Shell \MakeLowercase{\textit{et al.}}: A Sample Article Using IEEEtran.cls for IEEE Journals}


\maketitle

\begin{abstract}
Genetic algorithms (GA) have played an important role in engineering optimization. 
Traditional GAs treat each gene separately. 
However, biophysical studies of gene regulatory networks revealed direct associations between different genes. 
It inspires us to propose an improvement to GA in this paper, Gene Regulatory Genetic Algorithm (GRGA), which, to our best knowledge, is the first time to utilize relationships among genes for improving GA’s accuracy and efficiency. 
We design a directed multipartite graph encapsulating the solution space, called RGGR, where each node corresponds to a gene in the solution and the edge represents the relationship between adjacent nodes. 
The edge’s weight reflects the relationship degree and is updated based on the idea that the edges’ weights in a complete chain as candidate solution with acceptable or unacceptable performance should be strengthened or reduced, respectively. 
The obtained RGGR is then employed to determine appropriate loci of crossover and mutation operators, thereby directing the evolutionary process toward faster and better convergence. 
We analyze and validate our proposed GRGA approach in a single-objective multimodal optimization problem, and further test it on three types of applications, including feature selection, text summarization, and dimensionality reduction. 
Results illustrate that our GARA is effective and promising.
\end{abstract}


\section{Introduction}
\IEEEPARstart{S}{ince} the appearance of the genetic algorithm (GA) \cite{holland1975adaptation}, it has provided great help for the optimization of many engineering problems, and the improvement of GA has always been an important research direction. 
There are two categories of improvement. The first category focuses on the key operations of GA, such as coding \cite{2007The}, population \cite{korejo2013multi}, selection \cite{reeves2002genetic}, crossover \cite{qian2011analysis}, and mutation \cite{pham2022genetic}, \cite{xie2022improved}. 
The second category tries to develop new frameworks, such as parallel genetic algorithms through the parallel iteration of individuals \cite{hong2014effective}, chaotic genetic algorithms by adding a chaotic system \cite{javidi2015chaos}, and hybrid genetic algorithms through combining with other types of optimization methods \cite{el2006hybrid}.

\begin{figure*}[t]
\centering
    \includegraphics[width=7in]{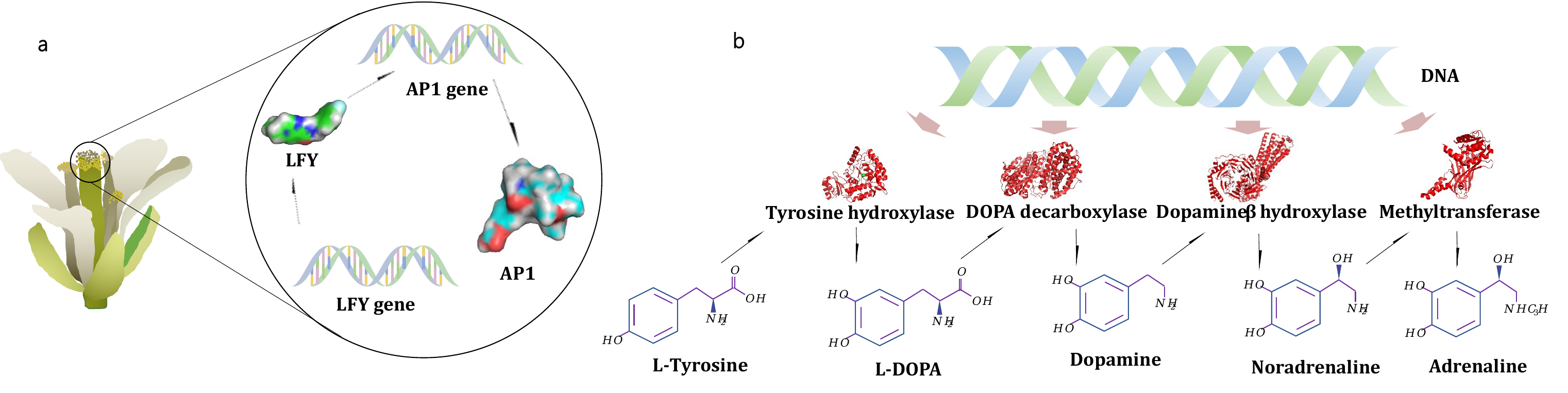}\\
    \caption{Interactions between genes. a. The expression product of the LEAFY (LFY) gene regulates APETALA1 (AP1) expression. b.The sequential process from tyrosine input to the final production of adrenaline.}
    \label{fig:fig1}
\end{figure*}

One shortcoming of GAs and their advanced extensions is the lack of consideration for the relationships among different solution components, i.e., each gene in a solution is independent. 
This is reflected in the fact that the loci of crossover and mutation operations are randomly selected, which could incur the unreasonable disruption of local interactive gene sequences, reducing the effectiveness of gene search \cite{1994Adaptive}. 
In fact, biological gene segments do not work independently of each other; instead, they interact with each other in various ways \cite{remm2001automatic}. 
We illustrate two cases in Fig. \ref{fig:fig1}(a). The first case is that one gene could regulate the expression of another. 
As shown in Fig. \ref{fig:fig1}(a), in Arabidopsis floral organ development, the expression product of the LEAFY (LFY) gene, which has the characteristics of a pioneer transcription factor, binds directly to chromatin condensation zones and alters the chromatin state of its target genes, regulating the expression of APETALA1 (AP1) \cite{lai2021leafy}. 
The second case is that the product of one gene expression requires the product of another gene expression to act before it can act. 
As shown in Fig. \ref{fig:fig1}(b), the process of the human body finally synthesizes adrenaline from tyrosine through the sequential process: L-Tyrosine → L-DOPA → Dopamine → Noradrenaline → Adrenaline, which suggests that there is a sequential effect of gene fragments on the whole regulatory process.

Drawing inspiration from gene regulatory networks in biophysics, this paper proposes a novel genetic algorithm through the consideration of sequential interactions of genetic codes, thereby enhancing performance and improving efficiency. 
We name this approach the Gene Regulatory Genetic Algorithm (GRGA). 
The core of GRGA is a directed Relationship Graph representing Gene Regulation (RGGR). 
Each directed chain in RGGR corresponds to a gene regulatory process. 
The larger the weight of an edge in the chain, the more reasonable the interaction between the two nodes is, and vice versa. 
Therefore, if the fitness of the individuals containing an edge tends to be promising, the weight of this edge will be gradually increased with the evolution process. 
In this way, the RGGR is updated. 
Then it is employed to guide gene crossover and mutation operations for reducing unreasonable disruption of interactive gene segments and finally improving the effectiveness and efficiency of searching. 
We evaluate the proposed GRGA on single-objective multimodal optimization problems and three applications including feature selection, text summarization, and dimension reduction, where GRGA is compared with the most advanced GAs.

The main contributions of this paper are listed as follows:
\begin{itemize}
    \item[(1)] To our best knowledge, this is the first genetic algorithm to consider the sequential interaction between genes to alleviate unreasonable disruption of local structures, which aligns with the mechanism of gene regulatory networks in biophysics.
    \item[(2)] We introduce a relationship graph representing gene regulation, called RGGR, and design a method to update RGGR in the evolution process for measuring interaction degrees between genes.
    \item[(3)] The RGGR is used to perform more reasonable crossover and mutation operations to improve the effectiveness and efficiency of searching.
    \item[(4)] The proposed GRGA can be widely applied to extend and improve various genetic algorithms. Careful experiments prove the advantages of our approach.
\end{itemize}

\section{Related Work}
\subsection{New frameworks of GA}
Our GRGA can be regarded as a new framework of genetic algorithms (GAs). 
Previous research in this direction mainly includes parallel, chaotic, and hybrid GAs. Parallel GAs can be further categorized into master-slave parallel, fine-grained parallel, and coarse-grained parallel \cite{hong2014effective}. 
The master-slave parallel GA is primarily used for parallel fitness calculation. 
The fine-grained parallel GA does not exchange individuals between subpopulations, and genetic operations are only performed on topologically adjacent subpopulations, while the coarse-grained parallel GA exchanges individuals between subpopulations. 
Chaotic GA addresses the premature convergence problem by introducing a chaotic system into the genetic algorithm \cite{javidi2015chaos}, where crossover and mutation operations are replaced by chaotic maps. 
Hybrid GAs \cite{el2006hybrid} combine GA with other optimization methods, such as Simulated Annealing (SA) \cite{islam2022hybrid}, Grey Wolf Optimizer (GWO) \cite{yadav2022hybrid}, and Tabu Search (TS) \cite{xu2021multi}, to improve performance.

In summary, these frameworks primarily incorporate genetic algorithms alongside other optimization techniques, rather than fundamentally altering the genetic algorithm itself.

\subsection{Crossover and mutation operators}
The intuitive role of our GRGA in improving traditional GAs is that the crossover and mutation loci are replaced by new ones from GRGA. 
Determining appropriate crossover and mutation positions by GRGA helps avoid unnecessary search space and concentrates the search in regions that may contain better solutions, thereby enhancing the search capability of genetic algorithms. 
Recent efforts in designing new crossover and mutation operators are briefly introduced below. 
Xue \cite{xue2021adaptive} used an adaptive operator selection mechanism and five crossover operators with different search characteristics to improve the performance of GAs in feature selection tasks. 
Hvattum \cite{hvattum2022adjusting} proposed an improved operator based on the OX crossover operator for hybrid genetic search, and enhanced the efficiency of the algorithm on the capacitated vehicle routing problem by filling the remaining part of the new solution into a specific location. 
Manzoni \cite{manzoni2020balanced} proposed and studied a variety of balanced crossover and recombination operators that maintain the Hamming weight of bit strings to improve the effectiveness of GAs in combination optimization problems. 
Yi \cite{yi2018improved} introduced an adaptive mutation operator to improve the performance of the standard NSGA-III algorithm. 
By designing a variable mutation rate, they improved its performance on multi-objective optimization problems.

Although many improved crossover and mutation operators have been proposed, the problem of deciding reasonable loci of crossover and mutation remains unexplored.

\section{The Proposed Method}
\subsection{The principle of GRGA}

\begin{figure*}[t]
\centering
    \includegraphics[width=7in]{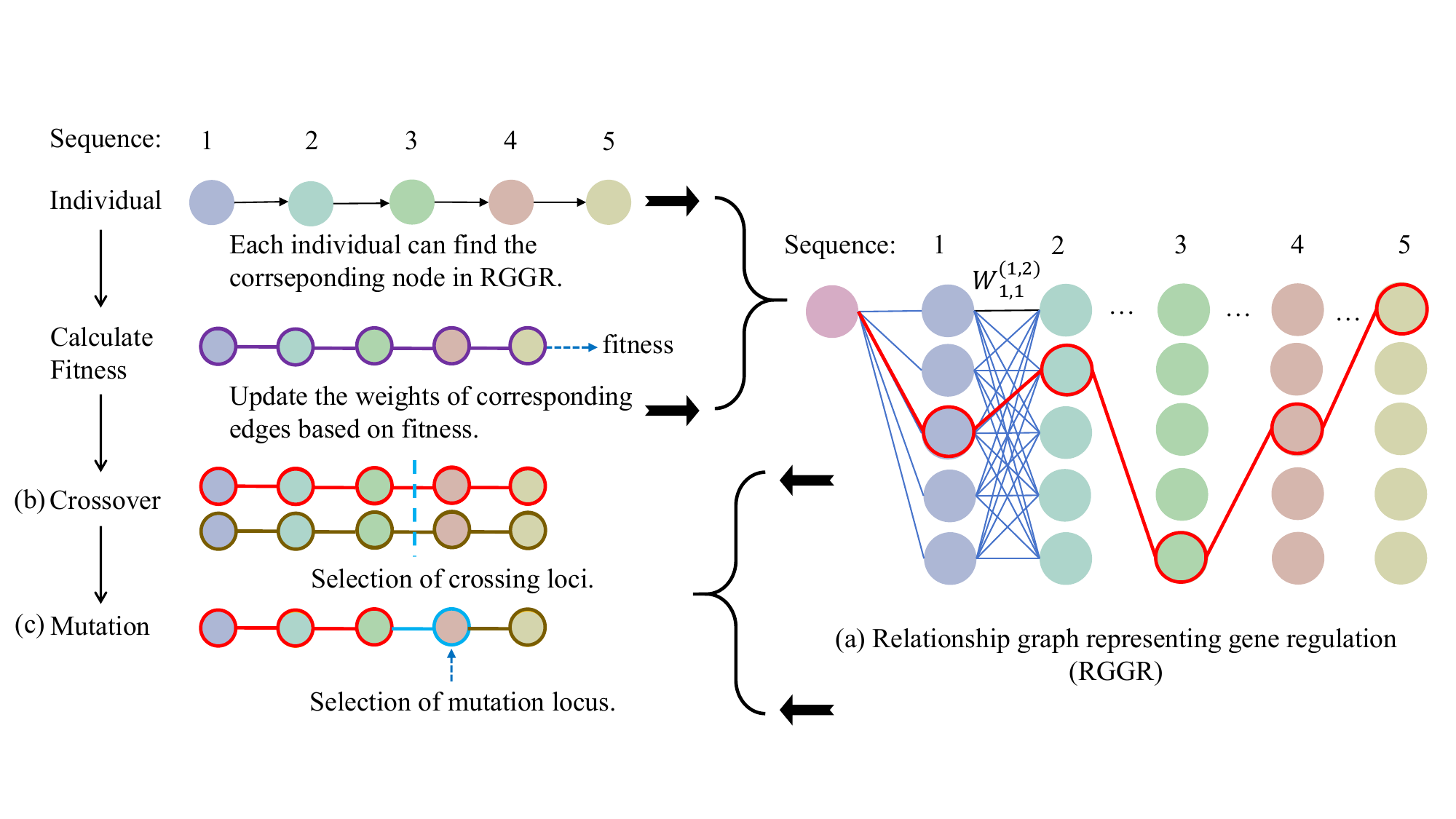}\\
    \caption{The working principle of our GRGA. a. Relationship graph representing gene regulation (RGGR), where the weight value in the edge represents the relationship degree between each part of the solution is updated from the fitnesses of individuals. b. Crossover loci were selected by RGGR. c. Mutation locus was selected by RGGR.}
    \label{fig:RGGR}
\end{figure*}

Let the alphabet set for coding the genes of a solution sequence to a problem be $\chi = \{1, 2, \ldots, n\}$.

Then a solution is represented as $G: 2-1-3-4-7$.

The working principle of our GRGA is illustrated in Fig. \ref{fig:RGGR} Taking the solution sequences with a length of 5 as an example, the solution space with the relationship between adjacent genes is represented by our RGGR as shown in Fig. \ref{fig:RGGR}(a)). 
Based on RGGR, the appropriate loci in the solution sequence for performing crossover and mutation are inferred in Fig. \ref{fig:RGGR}(b) and Fig. \ref{fig:RGGR}(c). Details are given in the following subsections.

\subsection{RGGR Calculation}

In RGGR, each node corresponds to a gene in the solution sequence, and the edges represent the relationship between two adjacent nodes.
We can analogize the process of searching for a solution to a gene regulatory pathway, in which each node unfolds in turn, with one node triggering the next node to occur.
The complete chain in RGGR is a single solution.
To simplify the problem, we assume a first-order Markov process, i.e., a node is only affected by its previous node.
The weight of an edge reflects the transition probability between adjacent nodes. 
We dynamically compare and adjust the transition probabilities between genes to guide the population to evolve toward a more favorable solution.

Let $n_i^k$ and $n_j^{k+1}$ represent the $i$-th node in the $k$-th column and the $j$-th node in the $(k+1)$-th column in RGGR, respectively. Let $W_{(i,j)}^{(k,k+1)}$ denote the weight of the edge between these two nodes, where $W_{(i,j)}^{(k,k+1)} \geq 0$. The larger $W_{(i,j)}^{(k,k+1)}$ is, the stronger the correlation between $n_i^k$ and $n_j^{k+1}$ is.

Initially, all $W_{(i,j)}^{(k,k+1)}$ are set to 1. 
They are then updated based on the quality of candidate solutions.
To do this, we introduce a threshold value $\lambda$ to determine whether a solution is acceptable. 
Let $\Delta$
represent the difference between the fitness value of the considered individual and the average fitness value of the current population.

If $\Delta \geq \lambda$, the individual as a complete chain in RGGR, exhibits sufficiently desirable performance. 
Consequently, each component of the solution is considered reasonable. 
Therefore, we can reinforce the weights of each local structure within the solution, facilitating more precise guidance for the genetic algorithm to alleviate unreasonable structures in subsequent iterations. 
This implies that the weights of the corresponding edges appearing in this chain in the RGGR should be strengthened. 
Formally,

\begin{equation}
\begin{cases}
W_{(i,j)}^{(k,k+1)} \neq 0 : \quad \quad \quad \text{(1-1)} \\
\quad W_{(i,j)}^{(k,k+1)} = \max\left\{W_{(i,j)}^{(k,k+1)} + \frac{\Delta}{\Delta + \rho} \cdot V\left(W_{(i,j)}^{(k,k+1)}\right),0\right\} ; \\
W_{(i,j)}^{(k,k+1)} = 0 : \quad \quad \quad \text{(1-2)} \\
\quad W_{(i,j)}^{(k,k+1)} = W_{(i,j)}^{(k,k+1)} + \mu\Delta
\end{cases}
\label{eq:e1}
\end{equation}


In Eq. 1, $V(W_{(i,j)}^{(k,k+1)}) > 0$ serves as a control function to prevent a certain $W_{(i,j)}^{(k,k+1)}$ from increasing excessively before receiving sufficient updates. 
The function $V$ is a function that varies with the problem and can also be a constant.
It needs to be designed according to specific applications. 
We provide examples in the experiment section. $\rho > 0$ is a control coefficient for the update rate. 
By introducing $\rho$, the influence of fitness $\Delta$ on the weight update can be limited within the range [0, 1], and $\mu$ within the range $(0,1)$.


In order to enhance the diversity of the population and improve the global search capability, we propose a fixed strategy to increase the competitiveness of nodes with $W_{(i,j)}^{(k,k+1)} = 0$. This strategy promotes the participation of these nodes in reproduction and gene exchange, thus enhancing population diversity and global search capability.

If $\Delta < \lambda$, indicating that there are unreasonable local structures in the solution sequence, the weight of the corresponding edges appearing in this chain in the RGGR can be reduced. 

Reversing Eq. 1, we obtain

\begin{equation}
\label{eq:e2}
\begin{cases}
W_{(i,j)}^{(k,k+1)} \neq 0 : \quad \quad \quad \text{(2-1)} \\
\quad W_{(i,j)}^{(k,k+1)} = \max\left\{W_{(i,j)}^{(k,k+1)} - \frac{\Delta}{\Delta + \rho} \cdot V\left(W_{(i,j)}^{(k,k+1)}\right),0\right\}; \\
W_{(i,j)}^{(k,k+1)} = 0 : \quad \quad \quad \text{(2-2)} \\
\quad W_{(i,j)}^{(k,k+1)} = \max\left\{W_{(i,j)}^{(k,k+1)} - \mu\Delta, 0\right\},
\end{cases}
\end{equation}

When an edge appears in multiple individuals simultaneously, its weight undergoes repeated adjustments using Eq. 1 and Eq. 2.

To make it less confusing, for the rest of this section, we use $W_n^{(k,k+1)}$ instead of $W_{n,(i,j)}^{(k,k+1)}$, $n$ stands for the $n-th$ individual share two same genetic loci.

\begin{figure}[ht]
\centering
    \includegraphics[width=3.5in]{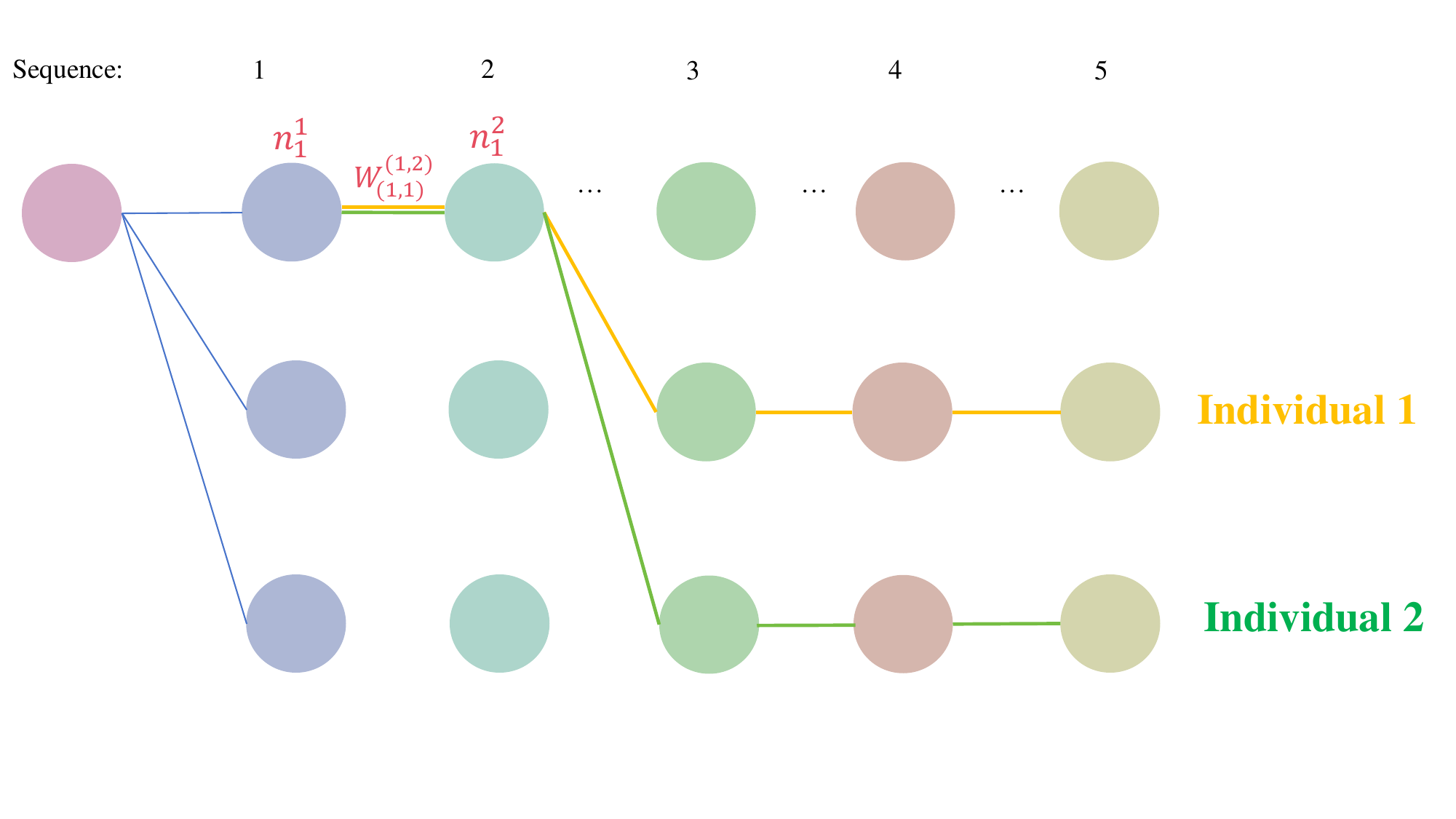}\\
    \caption{We give an example showing that when two gene loci are shared by two individuals, this is how the edges connected by these two individuals calculate the value of W.}
    \label{fig:fig_W}
\end{figure}

As shown in Fig. \ref{fig:fig_W}, we have two individuals, ``Individual 1 and Individual 2'', which individuals possess two loci, $n_1^1$ and $n_1^2$. $n_1^1$ and $n_1^2$ are connected by $W_{(1,1)}^{(1,2)}$, so this $W_{(1,1)}^{(1,2)}$ has to be updated twice: \\
$ W_1^{(1,2)} = W_{(1,1)}^{(1,2)} + \frac{\Delta}{\Delta + \rho} V(W_{(1,1)}^{(1,2)})$ \\
$ W_2^{(1,2)} = W_1^{(1,2)} + \frac{\Delta}{\Delta + \rho} V(W_1^{(1,2)}$

We define that when an edge is shared by N individuals in the population(Eq. \ref{eq:eq2}), the weight $W_n^{(k,k+1)}$ requires recalculation and updating, resulting in N updates corresponding to the number of individuals sharing the edge.

\begin{equation}
\label{eq:eq2}
\resizebox{0.9\linewidth}{!}{$
W_{n+1}^{(k,k+1)} = W_{(i,j)}^{(k,k+1)} + \sum_{n=1}^{N}\left(\frac{\Delta}{\Delta + \rho} \cdot V\left(W_{n}^{(k,k+1)}\right)\right)
$}
\end{equation}
When $n = 0$, $W_{0}^{(k,k+1)} = W_{(i,j)}^{(k,k+1)}$,When $n = 0$, $W_{0}^{(k,k+1)} = W_{(i,j)}^{(k,k+1)}$ implies that no two or more individuals share two same genetic loci.

These adjustments occur separately for each instance of the edge in different individuals, leading to multiple updates. 
This approach helps maintain population diversity and enhances the algorithm's exploration of the solution space, reducing the chances of encountering local optima.

\begin{figure*}[t]
\centering
    \includegraphics[width=7in]{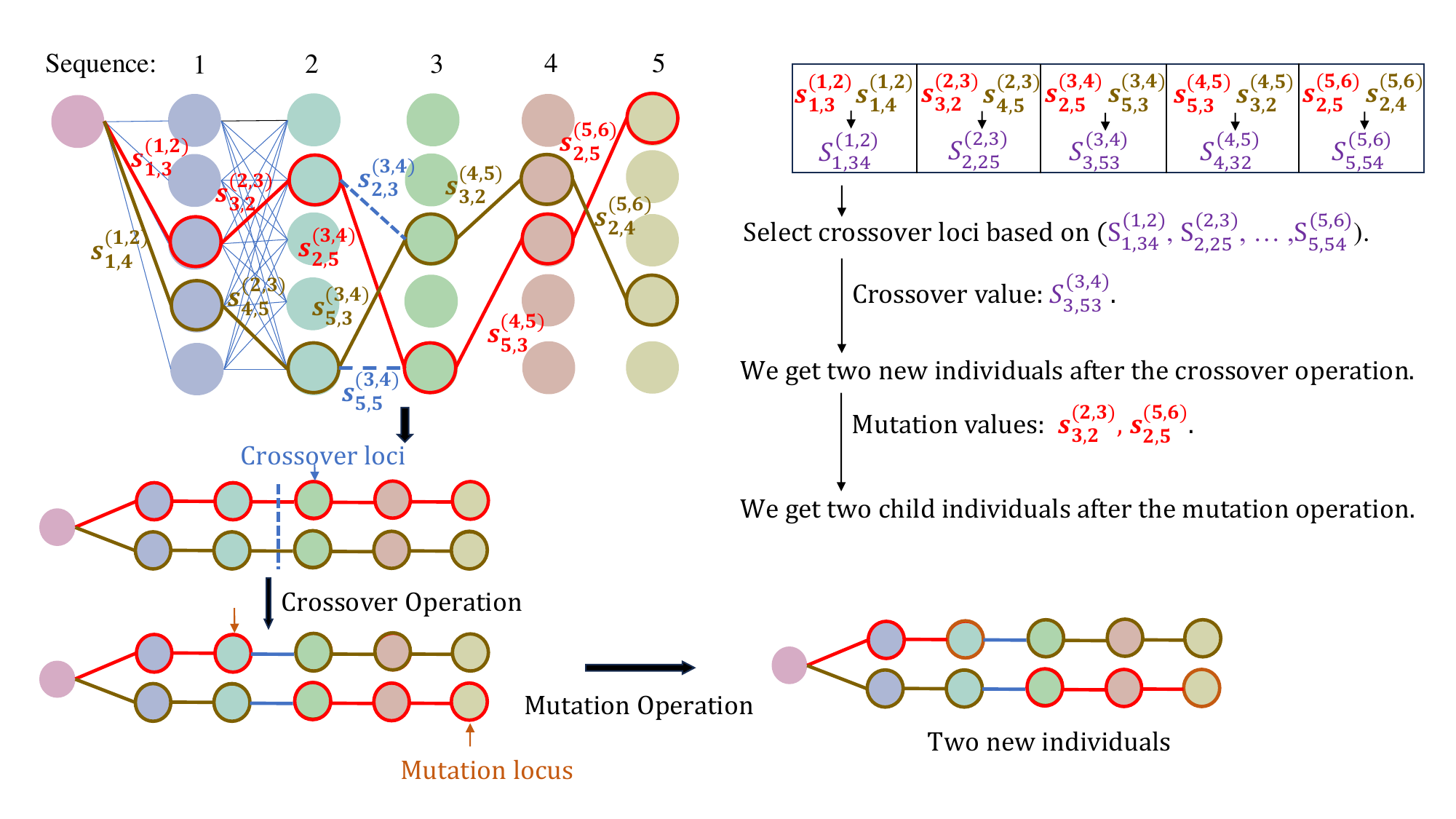}\\
    \caption{How to choose crossover and mutation loci. We provide an example to illustrate the crossover and mutation processes of GRGA in detail.}
    \label{fig:example}
\end{figure*}

\subsection{Inferring Crossover Loci and Mutation Locus based on RGGR}
After the RGGR is updated as above, we use it to guide the crossover and mutation operators. 
Reasonable crossover and mutation loci can make the offspring evolve towards the optimal solution more accurately. 
We calculate the following formula

\begin{equation}
s_{(i,j)}^{(k,k+1)} = \frac{1}{C_1 + C_2 W_{(i,j)}^{(k,k+1)}}
\label{eq:e3}
\end{equation}

as the basis of selecting crossover and mutation loci. 
In Eq. \ref{eq:e3}, the introduction of $C_1$ and $C_2$ in the equation is intended to regulate the effect of $W_{(i,j)}^{(k,k+1)}$. $C_1$ ensures that calculations can be performed even if $W_{(i,j)}^{(k,k+1)} = 0$, 
while $C_2$ serves to adjust the influence of $W_{(i,j)}^{(k,k+1)}$. 
The crossover process initiates with the computation of the selection probability for each gene locus to potentially act as a crossover locus. 
This computation relies on the strength values attributed to the genes within the individuals' genomes. 
We calculate the sum of $s_{i,j}^{(k,k+1)}$ for alleles of the two parent chains, like $S_{i,j_1j_2}^{(k,k+1)} = s_{i,j_1}^{(k,k+1)} + s_{i,j_2}^{(k,k+1)}$.

As shown in Fig. \ref{fig:example}, we compute $S_{i,j_1j_2}^{(k,k+1)}$, such as \(  S_{1,34}^{(1,2)} =  s_{1,3}^{(1,2)} +  s_{1,4}^{(1,2)} \). 

Suppose the results are $s_{i,j}^{(k,k+1)}$ and $S_{i,j_1j_2}^{(k,k+1)}$, based on which we calculate the probability $p_k$ of selecting the $k-th$ locus for performing crossover and mutation by Eq. \ref{Eq:e4} and Eq. \ref{Eq:e42}, respectively, where $K = length$ $of$ $gene$ $sequence-1$ and is the number of possible loci.
\begin{equation}
P_{crossover} = \frac{S_{i,j_1j_2}^{(k,k+1)}}{\sum_{k=0}^{K} S_{i,j_1j_2}^{(k,k+1)}} \times 100\%
\label{Eq:e4}
\end{equation}

\begin{equation}
P_{mutation} = \frac{s_{i,j}^{(k,k+1)}}{\sum_{k=0}^{K} s_{i,j}^{(k,k+1)}} \times 100\%
\label{Eq:e42}
\end{equation}


To sum up, we give the pseudo code of our GRGA in Algorithm \ref{alg:GRGA}.

\begin{algorithm}
\caption{Gene Regulatory Genetic Algorithm (GRGA)}
\label{alg:GRGA}
\begin{algorithmic}[1]
\State $Parents \gets$ Random initialization population
\State $Child \gets \emptyset$
\While{not (Termination conditions)}
    \State Gene sequence expressed as genetic representation
    \State Generate the solution from the representation, calculate the effectiveness of the solution as the fitness
    \State Update the genetic representation action diagram according to Eq. 1 and Eq. 2 above: $W_{(i,j)}^{(k,k+1)}$
    \State Update the new RGGR according to Eq. \ref{eq:e3}
    \State Dynamically adjust the probability of selecting crossover loci by balancing the relationship between individual similarities and maintaining population diversity as described in Eq. \ref{Eq:e4}.
    \State Dynamically adjust the probability of selecting mutation locus by considering individual mutation frequencies as described in Eq. \ref{Eq:e42}.
    \State $Child \gets$ survival selection
\EndWhile
\State \Return solution
\end{algorithmic}
\end{algorithm}

\subsection{Apply the proposed method}
GRGA does not require specific selection, crossover, or mutation operations, we can easily extend traditional GAs into GRGA strengthened ones as follows:
\begin{itemize}
    \item Design the RGGR according to the search space of the problem.

    \item After obtaining the fitness of individuals, update RGGR according to the method described in Section 3.B.
    
    \item In crossover and mutation operations, replace randomly selected loci with RGGR selected loci.
\end{itemize}

\section{Experiments}
We first validate and analyze our GRGA by conducting function evaluations using a benchmark from Competitive Evolutionary Computation (CEC), 
then we test it in three typical applications of genetic algorithm, 
including feature selection, text summarization, and dimension reduction.
The proposed GRGA is embedded into the latest and most advanced GAs for these three applications. 

\subsection{Verification and Analysis of GRGA}
We opt for the Shubert(3 dimensions) function from the CEC2013 test function set as our experimental subject. 
This choice stems from the function's inherent multimodal characteristics, which present challenges in navigating intricate solution spaces with multiple peaks.

The expression for Shubert is given by:

\begin{equation}
F(x) = -\prod_{i=1}^{D} \sum_{j=1}^{5} j \cos[(j+1) x_i + j]
\label{Eq:4}
\end{equation}

This function has a total of 81 peaks, with the maximum value being 2709.0935. 
We configured the parameters of GRGA with a population size of 200, mutation rate of 0.05, evolution generations of 30, and a scaling factor (f) of 0.3. 
We conducted a comprehensive analysis through 100 Monte Carlo experiments. 
To ensure consistency with the comparison, we set GA parameters identical to those of GRGA.

We aim to rigorously assess GRGA's effectiveness in simultaneously exploring and converging to diverse peaks, highlighting its robustness and adaptability in handling complex optimization landscapes.

\begin{figure*}[t]
    \includegraphics[width=7.5in]{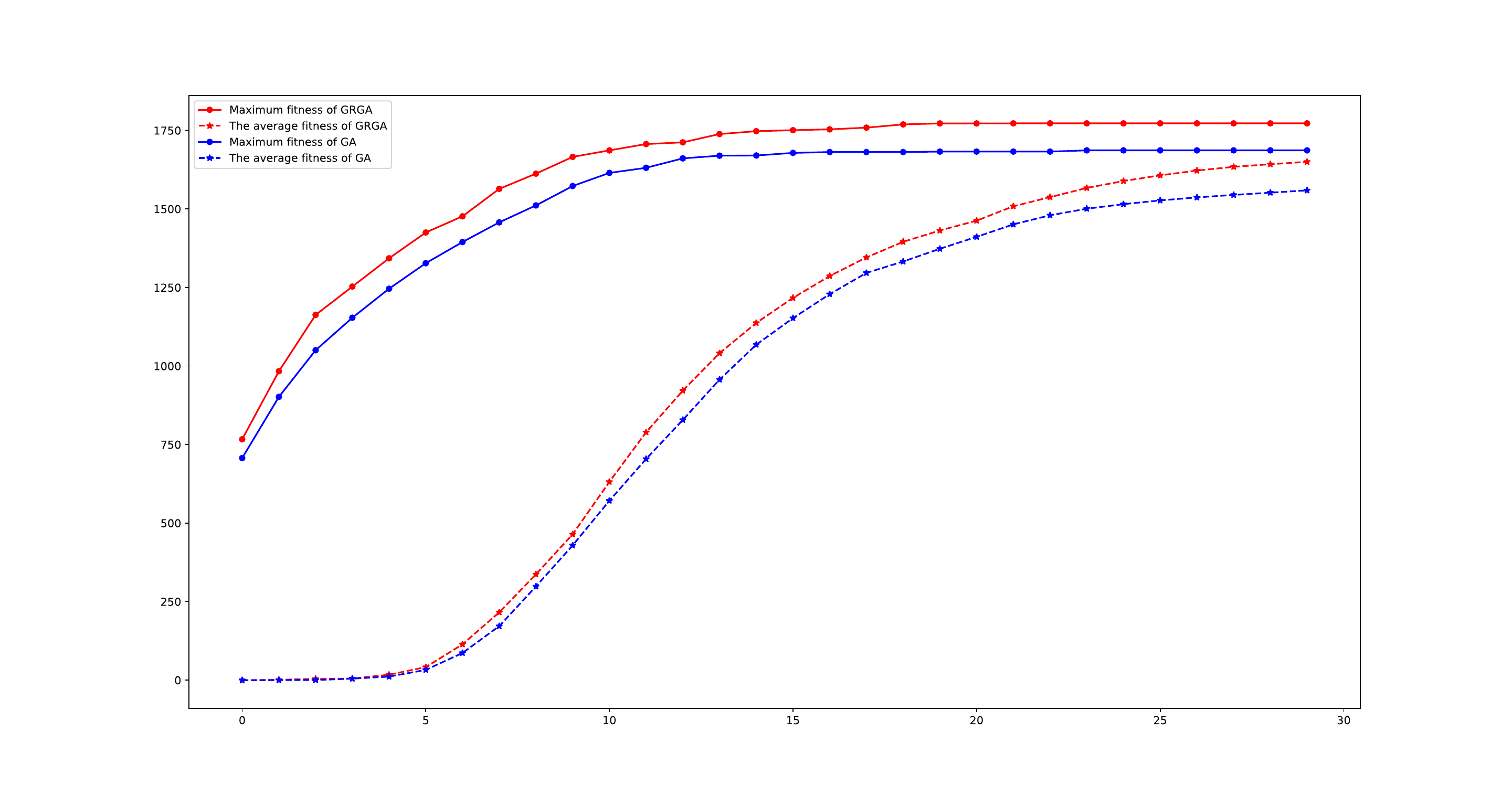}\\
    \caption{The evolution of maximum and average fitness of GRGA and original GA over 100 experiments with evolving generations.}
    \label{fig:GAGRGA}
\end{figure*}

Fig. \ref{fig:GAGRGA} illustrates the evolution of average maximum fitness and population average fitness over 100 experiments with evolving generations. 
The results showed that the difference between GRGA and original GA was not significant in the first 5 generations of population average fitness. 
However, after the 10th generation, the population average fitness of GRGA began to significantly exceed that of the original GA. 
GRGA's ascent rate and final convergence performance surpass those of the original GA.

\begin{figure}[t]
\centering
    \includegraphics[width=3.9in]{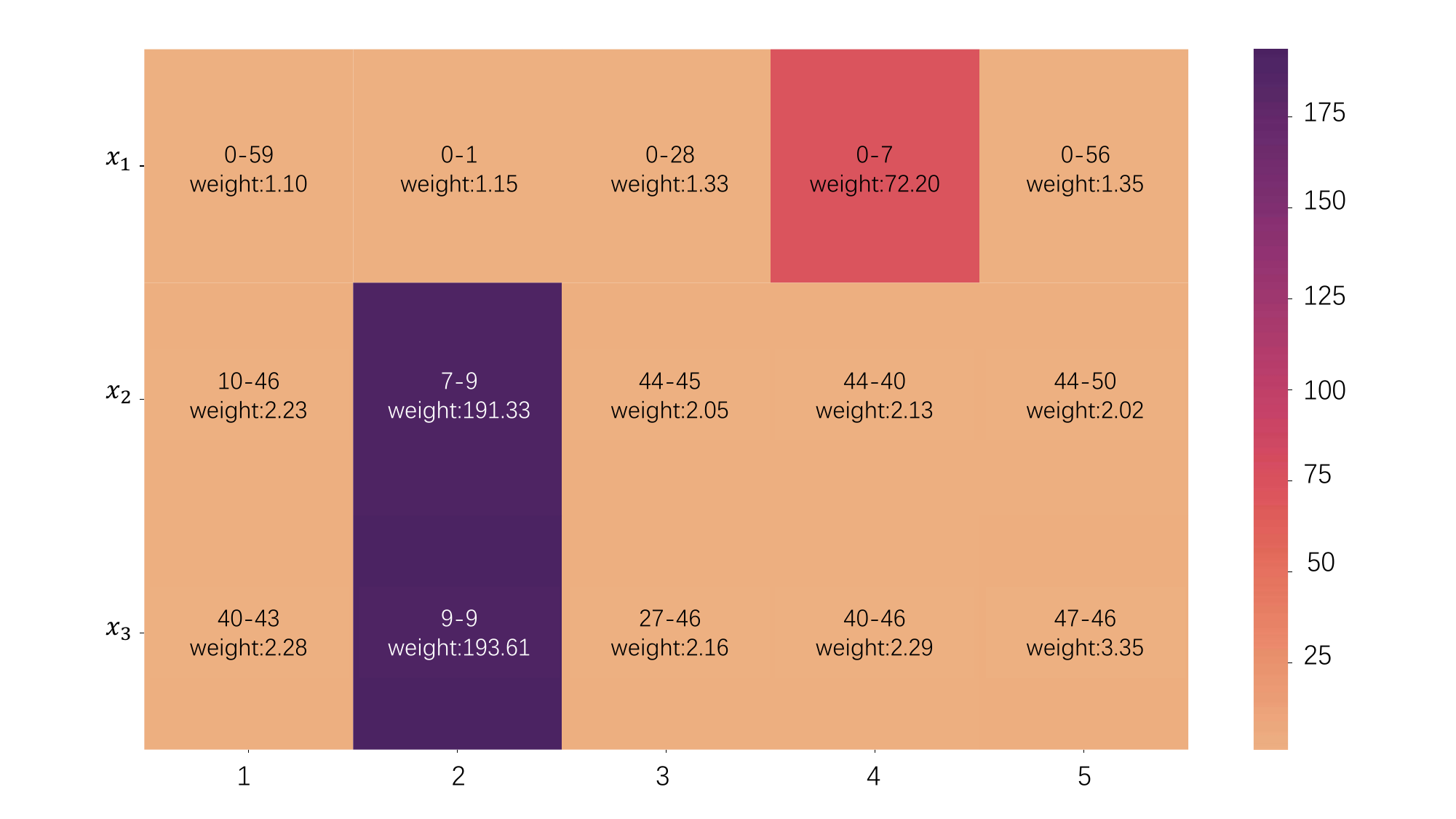}\\
    \caption{ TOP5 weighted heatmap. Each row represents a dimension of the solution space. In the figure, the colors indicate the top five genetic loci with the highest weights in each dimension. The depth of color reflects the magnitude of the weight value, where a darker color signifies a larger weight value. }
    \label{fig:fig6}
\end{figure}

As the RGGR structure in GRGA currently only applies to discrete genes, we discretized the three variables of Shubert function within the interval [-10,10] into 60 integers from 0 to 59. 
The values of nodes in the RGGR displayed in Fig. \ref{fig:fig6} correspond to these discrete values.
In RGGR, determine the top five most weighted values for each gene sequence in the last generation.

As depicted in Fig. \ref{fig:fig6}, the peak weight for $x_1$ is observed at the 7th interval, with a value of 72.20. 
Considering the function's solution space spans from $(-10, 10)$, and each dimension is subdivided into 60 intervals, the computation of $x_1$ is performed as follows:
\[ W_{0,7}^{(0,1)}: x_1 = \left(\frac{[10-(-10)] \times 7}{60}\right) + (-10) = \frac{-23}{3}; \]
Similarly, we can derive that:

$\max(W_{(i,j)}^{(1,2)}) = W_{7,9}^{(1,2)} = 191.33$, 
\[ W_{7,9}^{(1,2)}: x_2 = \left(\frac{[10-(-10)] \times 9}{60}\right) + (-10) = -7; \]

\begin{figure*}[t]
\centering
    \includegraphics[width=7in]{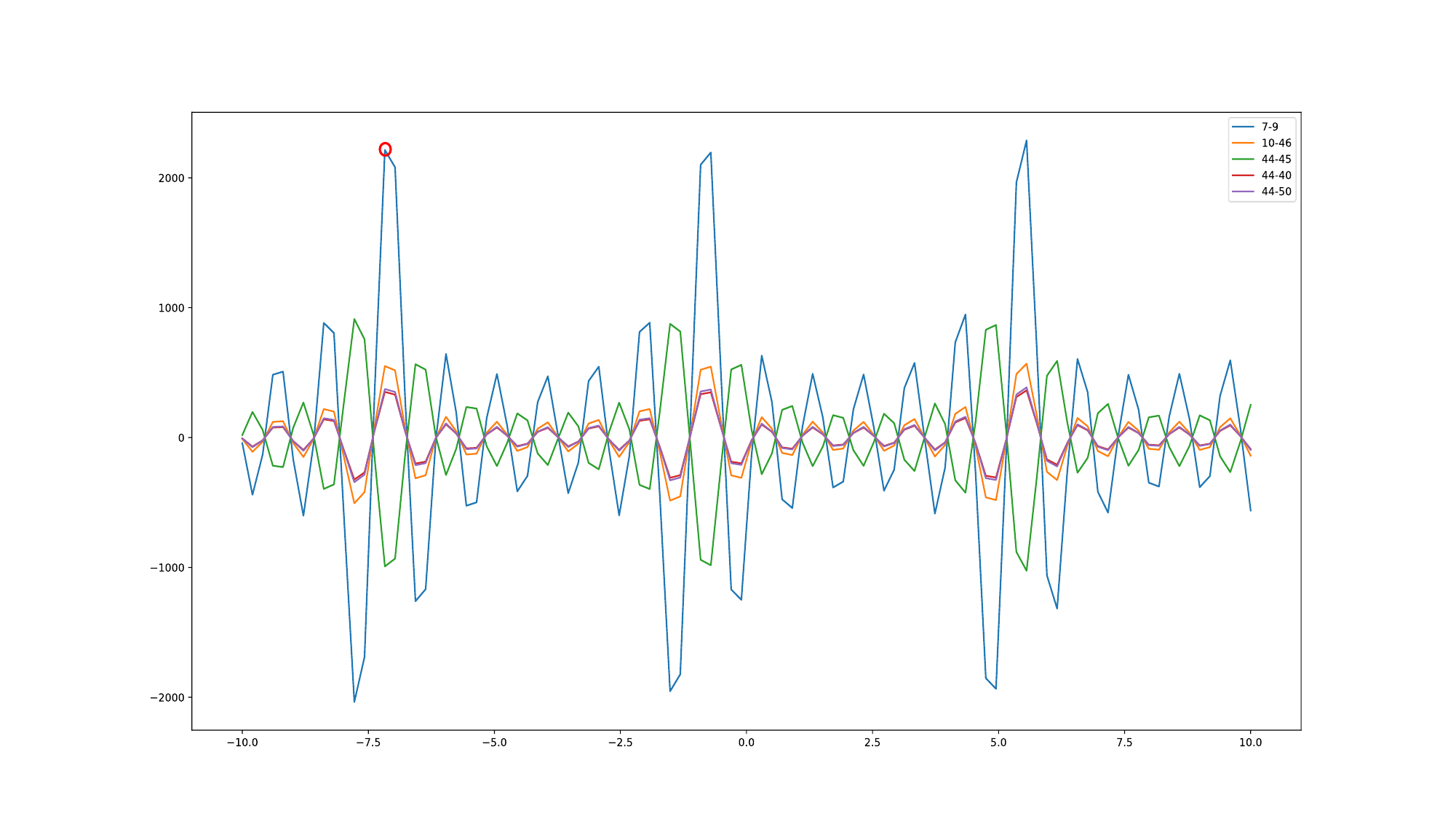}\\
    \caption{After fixing the combination relationship of $x_1$ and $x_2$, the impact of the value of $x_3$ on the fitness of the function. The independent variable value represented by the structure with the largest weight can obtain a higher peak value. The notation $x_1$-$x_2$ in the legend represents $x_1$ taken from the $n_1$-th interval in the first dimension of the solution space, and $x_2$ taken from the $n_2$-th interval in the second dimension. The curves in the graph illustrate the variation of $x_3$ across the third dimension of the solution space, corresponding to changes in the Shubert function's values.}
    \label{fig:fig5}
\end{figure*}

$\max(W_{(i,j)}^{(2,3)}) = W_{9,9}^{(2,3)} = 193.61$. 
\[ W_{9,9}^{(2,3)}: x_3 = \left(\frac{[10-(-10)] \times 9}{60}\right) + (-10) = -7. \]
The points($\frac{-23}{3},-7,-7$), which as shown in the red circle in Fig. \ref{fig:fig5} represent the global optimal solutions found by GRGA.

Through in-depth analysis of the experimental results, we focus on the key component of GRGA, i.e., RGGR, and reveal its function and role in the optimization process.

\paragraph{Local structure guidance}

In RGGR, the weight reflects the quality of each local structure, indicating the potential effectiveness of specific combinations of independent variables.
As depicted in Fig. \ref{fig:fig5}, each line represents the fitness function's variation with $x_1$ and $x_2$ held constant, while $x_3$ varies within the range of -10 to 10. 
It is notable that only the blue line corresponds to the instance where the maximum fitness value is achieved. 
This observation underscores RGGR's capacity to autonomously select local structures, guiding the overall evolution process.

\paragraph{Global optimization capability}

In the experiment, RGGR enables GRGA to rapidly and effectively search for and converge to optimal solutions within fewer generations. 
Specifically, the combination $x_1 = \frac{-23}{3}$, $x_2 = -7$, and $x_3 = -7$ emerged as optimal. 
By dynamically adjusting the weights of local structures, RGGR enhances GA's search efficiency, facilitating quicker convergence towards global optimal solutions.

\paragraph{Interaction between independent variables}

Fig. \ref{fig:fig5} vividly illustrates the impact of $x_3$ on the function value, emphasizing the interplay between independent variables. 
Notably, the selection of $x_1$ and $x_2$ significantly influences achieving higher function values. 
This showcases RGGR's capability to capture intricate relationships among independent variables, thereby enhancing the likelihood of discovering potential global optimal solutions.

Overall, RGGR plays a key role in self-selected crossover and mutation operations through its weight structure. 
This allows GRGA to selectively emphasize the self-selection of structures with higher potential fitness, thereby achieving a more efficient global search throughout the optimization process. 

\subsection{Application Evaluation}
In three application tests, the GRGA parameters are set as follows based on experiments. 
In Eq. \label{eq:e1}1-1 and Eq. \label{eq:e2}2-1, $\rho$ is set to 0.1 of the average fitness of the contemporary population. 
$\mu$ in Eq. \label{eq:e1}1-2 and Eq. \label{eq:2}2-2 is set to 0.8. $C_1$ and $C_2$ in Eq. \ref{eq:e3} are taken as 1 and 0.1, respectively. 
For Eq. \label{eq:e1}1-1 and Eq. \label{eq:e2}2-1, if $k=0$, the update value is multiplied by 0.5. 
The termination condition is set such that the best fitness of the population remains unchanged for ten consecutive generations.

\subsubsection{Feature Selection}
\paragraph{Problem Description}
To address the feature selection problem, Altarabichi et al \cite{ALTARABICHI2023118528} proposed a new fast genetic algorithm (CHCqx). 
To verify the advantages of our algorithm in feature selection, we conducted experiments with our algorithm (GRGA-CHCqx) on this problem and compared it with CHCqx.

\paragraph{Result Analysis}
We recorded the running time of the original method and the method extended by our GRGA. 
The results are presented in Table \ref{table:CHCqx1}. 
The average running time of the original method and our extended method is 39.92 seconds and 27.07 seconds, respectively. 
Efficiency has increased by more than 32.19\%. 
Additionally, the effectiveness of feature selection has slightly improved, as shown in Table \ref{table:CHCqx1}.

\begin{table*}[hbtp]
\centering
\caption{The running time of GAs in feature selection.}
\label{table:CHCqx1}
\begin{tabular}{|l|*{11}{c}|}
\hline
 & 1 & 2 & 3 & 4 & 5 & 6 & 7 & 8 & 9 & 10 & mean \\
\hline
CHCqx & 45.13 & 39.78 & 34.19 & 30.08 & 30.99 & 28.11 & 32.36 & 69.98 & 43.25 & 45.36 & 39.92 \\
GRGA-CHCqx & 23.12 & 32.35 & 37.28 & 29.59 & 16.68 & 28.74 & 31.13 & 16.49 & 27.5 & 27.8 & 27.07 \\
\hline
\end{tabular}
\end{table*}

\begin{table*}[hbtp]
\centering
\caption{The score (accuracy) of feature selection.}
\label{table:CHCqx2}
\begin{tabular}{|l|*{11}{c}|}
\hline
& 1 & 2 & 3 & 4 & 5 & 6 & 7 & 8 & 9 & 10 & Mean \\
\hline
CHCqx & 94.93 & 94.81 & 94.93 & 94.9 & 94.94 & 94.94 & 94.94 & 94.94 & 94.9 & 94.94 & 94.917 \\
GRGA-CHCqx & 94.92 & 94.89 & 94.94 & 94.95 & 94.95 & 94.92 & 94.92 & 94.93 & 94.94 & 94.94 & 94.93 \\
\hline
\end{tabular}
\end{table*}

\begin{table}[hbtp]
\centering
\caption{Operation Time (s)}
\label{table:OPTime}
\resizebox{0.45\textwidth}{!}{
\begin{tabular}{|l*{2}{c}|}
\hline
& DUC2005 & DUC2007 \\
\hline
MTSQIGA & 278.75 & 123.48 \\
 & 283.96 & 116.83 \\
 & 282.61 & 126.22 \\
\hline
mean & 281.7733 & 122.1767 \\
\hline
GRGA-MTSQIGA & 277.98 & 110.14 \\
 & 283.4 & 124.25 \\
 & 287.91 & 123.18 \\
\hline
mean & 283.0967 & 119.19 \\
\hline
\end{tabular}
}
\end{table}

\subsubsection{Text Summary}
\paragraph{Problem Description}
In the task of automatically generating text summaries, the quantum heuristic genetic algorithm (MTSQIGA) \cite{mojrian2021novel} has achieved good results on DUC 2005 and 2007 benchmark datasets. 
We conducted experiments with our algorithm (GRGA-MTSQIGA) on this problem and compared it with the original method.

\begin{table*}[hbtp]
\centering
\caption{ROUGE scores on DUC 2005.}
\label{table:MTSQIGA1}
\begin{tabular}{|l|ccc|ccc|}
\hline
& \multicolumn{3}{c|}{MTSQIGA} & \multicolumn{3}{c|}{GRGA-MTSQIGA} \\
\cline{2-4} \cline{5-7}
& ROUGE-1 & ROUGE-2 & ROUGE-SU4 & ROUGE-1 & ROUGE-2 & ROUGE-SU4 \\
\hline
Average F-Score & 0.354247 & 0.08112 & 0.13746033 & 0.3595 & 0.081011 & 0.141524 \\
Average Recall & 0.349966 & 0.080165 & 0.136083 & 0.355823 & 0.080181 & 0.139995333 \\
Average Precision & 0.358736 & 0.082108 & 0.13889133 & 0.363382 & 0.081873 & 0.14311 \\
\hline
\end{tabular}
\end{table*}

\begin{table*}[hbtp]
\centering
\caption{ROUGE scores on DUC 2007.}
\label{table:MTSQIGA2}
\begin{tabular}{|l|ccc|ccc|}
\hline
& \multicolumn{3}{c|}{MTSQIGA} & \multicolumn{3}{c|}{GRGA-MTSQIGA} \\
\cline{2-4} \cline{5-7}
& ROUGE-1 & ROUGE-2 & ROUGE-SU4 & ROUGE-1 & ROUGE-2 & ROUGE-SU4 \\
\hline
Average F-Score & 0.447365 & 0.120771 & 0.18251333 & 0.453382 & 0.124821 & 0.186347667 \\
Average Recall & 0.445665 & 0.120547 & 0.182943 & 0.453013 & 0.124667 & 0.187198667 \\
Average Precision & 0.449274 & 0.121048 & 0.182166 & 0.453983 & 0.125036 & 0.185615667 \\
\hline
\end{tabular}
\end{table*}

\begin{figure}[t]
\centering
\includegraphics[width=4.3in]{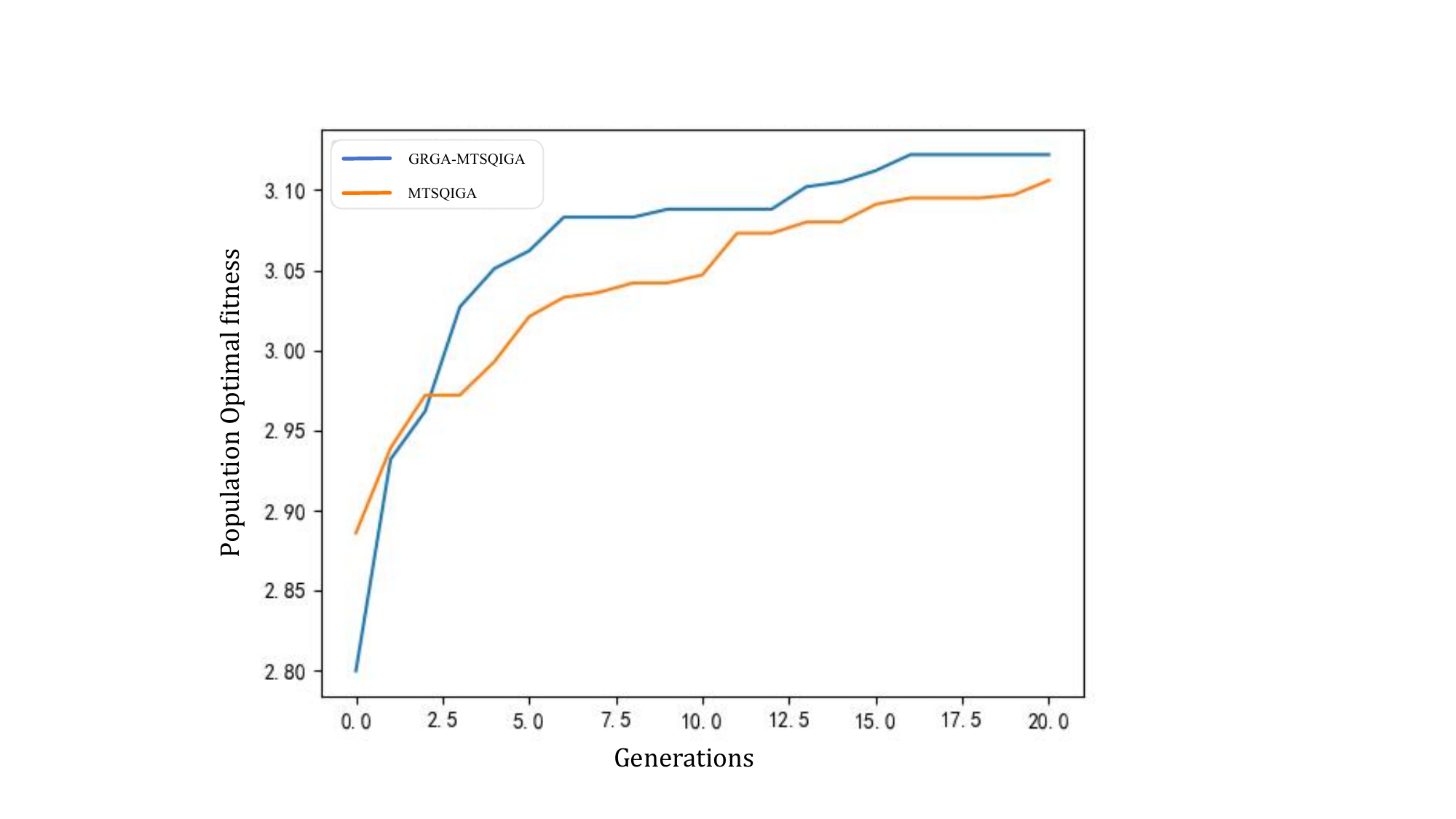}\\
\caption{ Result of accuracy changing with generations.}
\label{fig:fig7}
\end{figure}

\paragraph{Result Analysis}
We recorded the change of average in accuracy of 10 outputs on the data of d438g in DUC 2005. 
The results are shown in Fig. \ref{fig:fig7}. 
Given that the original algorithm may terminate at different generations, we selected the shortest duration, which comprised 21 generations, for comparison. 
This choice was based on our termination criterion, which stipulated that termination occurs when the fitness remains unchanged for ten consecutive generations.
It can be seen that the accuracy of GRGA-MTSQIGA improves faster than that of MTSQIGA. 
This indicates that RGGR provides a better search direction by guiding crossover and mutation, which improves the search efficiency of the algorithm. 
In terms of convergence effects, in F-score, recall, and precision, the GRGA-MTSQIGA is significantly better than the original method, as shown in Table \ref{table:MTSQIGA1} and Table \ref{table:MTSQIGA2}. 
We also tested the algorithm's computing efficiency. 
As shown in Table \ref{table:OPTime}, it can be observed that the addition of our approach has little effect on the overall computational speed of the algorithm. 
In addition, MTSQIGA adopts quantum encoding, which indicates that our algorithm is robust to the encoding form.

\subsubsection{Dimensionality Reduction}
\paragraph{Problem Description}
In the task of data dimensionality reduction, literature \cite{radeev2023transparent} provides a genetic algorithm GDR for transparent dimensionality reduction of numerical data. We incorporate our GRGA into the GDR algorithm (GDR\_GRGA).

\begin{table*}[hbtp]
\caption{Comparative experimental results in dimensionality reduction}
\centering
\resizebox{\textwidth}{!}{%
\begin{tabular}{lcccccc}
Dataset      & GDR PostF1 & GDR PostPrec & GDR PostRecall & GDR\_GRGA PostF1 & GDR\_GRGA PostPrec & GDR\_GRGA PostRecall \\ \hline
bank\_marketing     & 0.427      & 0.465        & 0.438          & 0.437             & 0.468               & 0.460                \\
blood\_trans        & 0.374      & 0.524        & 0.297          & 0.439             & 0.544               & 0.394                \\
Breast cancer      & 0.897      & 0.906        & 0.888          & 0.951             & 0.923               & 0.980                \\
Credit g           & 0.795      & 0.719        & 0.893          & 0.775             & 0.721               & 0.843                \\
bioresponse        & 0.352      & 0.638        & 0.374          & 0.741             & 0.745               & 0.737                \\
ionosphere         & 0.896      & 0.891        & 0.902          & 0.870             & 0.907               & 0.837                \\
sonar              & 0.734      & 0.728        & 0.744          & 0.775             & 0.718               & 0.844                \\
christine          & 0.610      & 0.609        & 0.619          & 0.637             & 0.644               & 0.630                \\
hyperplane         & 0.643      & 0.673        & 0.616          & 0.660             & 0.614               & 0.719                \\
diabetes           & 0.629      & 0.680        & 0.588          & 0.593             & 0.614               & 0.576                \\
madelon            & 0.640      & 0.635        & 0.646          & 0.665             & 0.652               & 0.679                \\
guillermo          & 0.437      & 0.615        & 0.369          & 0.397             & 0.536               & 0.342                \\ \hline
\end{tabular}%
}
\label{tab:dimensionality_reduction}
\end{table*}

\paragraph{Result Analysis}
We recorded the average of 5 Monte Carlo experiments as the final result to show in Table \ref{tab:dimensionality_reduction}. 
Firstly, in most issues, GDR\_GRGA outperforms GDR in F-score, precision, and recall. 
Especially in the bioesponse problem, GDR\_GRGA has achieved much better performance than the original algorithm. 
The cases in which GDR\_GRGA may not be suitable for higher-order Markov models, resulting in the performance of RGGR based on first-order Mar-kov processes is not good enough.
For the cases in which GDR\_GRGA's performance is not good enough, maybe we need to extend RGGR through cosidering higher-order Markov  chains.

\section{Conclusions}
Inspired by the gene regulatory networks in biophysics, this paper has proposed a Gene Regulatory Genetic Algorithm (GRGA) that models the relationship between genes and uses it to guide crossover and mutation operations, thereby improving the efficiency and effectiveness of GAs. 
We demonstrated the effectiveness of GRGA and the role of RGGR through a test function in CEC2013. 
GRGA further shows promising results in three applications. 
In the feature selection task, the speed of convergence increased 32.19\% than the most advanced GA for this problem while maintaining the accuracy. 
We increase average score by more than 0.5\% on text summary tasks. 
The effect is superior to state-of-the-art GA in over 66\% of dimensionality reduction tasks. 
We only consider the first-order Markov relationship in this work, introducing higher-order Markov relationship could further improve the performance of our approach, which will be explored in our future work.

\section*{Acknowledgments}
This should be a simple paragraph before the References to thank those individuals and institutions who have supported your work on this article.

%

\bibliographystyle{IEEEtran}
\bibliography{ref}





\end{document}